# Hybrid Adaptive Control for Series Elastic Actuator of Humanoid Robot

Anh Khoa Lanh Luu[1], Van Tu Duong[1,2], Huy Hung Nguyen[1,3], Sang Bong Kim[4] and Tan Tien Nguyen[1,*]

[1]*National Key Laboratory of Digital Control and System Engineering, Ho Chi Minh City University of Technology, VNU-HCM, Hochiminh city, Vietnam.*

[2]*Department of Mechatronics, Ho Chi Minh City University of Technology, VNU-HCM, Hochiminh city, Vietnam.*

[3]*Faculty of Electronics and Telecommunication, Saigon University, Vietnam*

[4]*Pukyong National University, Busan, Republic of Korea*

[*]Corresponding author. E-mail: *nttien@hcmut.edu.vn*

## Abstract

**Purpose -** Generally, humanoid robots usually suffer significant impact force when walking or running in a non-predefined environment that could easily damage the actuators due to high stiffness. In recent years, the usages of passive, compliant series elastic actuators (SEA) for driving humanoid's joints have proved the capability in many aspects so far. However, despite being widely applied in the biped robot research field, the stable control problem for a humanoid powered by the SEAs, especially in the walking process, is still a challenge. This paper proposes a model reference adaptive control (MRAC) combined with the back-stepping algorithm to deal with the parameter uncertainties in a humanoid's lower limb driven by the SEA system. This is also an extension of our previous research (Lanh *et al.*, 2021).

**Design/methodology/approach -** Firstly, a dynamic model of SEA is obtained. Secondly, since there are unknown and uncertain parameters in the SEA model, a model reference adaptive controller (MRAC) is employed to guarantee the robust performance of the humanoid's lower limb. Finally, an experiment is carried out to evaluate the effectiveness of the proposed controller and the SEA mechanism.

**Findings -** This paper proposes an effective control algorithm that can be widely applied for the humanoid-SEA system. Besides, the effect of the coefficients in the control law is analyzed to further improve the response's quality.

**Research limitations/implications -** Even though the simulation shows promising results with stable system response, the practical experiment has not been implemented to fully evaluate the quality of the controller.

**Originality/value -** The MRAC is applied to control the humanoid's lower limb, and the back-stepping process is utilized to combine with an external SEA system but still maintain stabilization. The simplified model of the lower-limb system proposed in the paper is proven to be appropriate and can be taken for further research in the future.

**Keywords** Series elastic actuators (SEAs), humanoid robot, back-stepping algorithm, model reference adaptive control (MRAC).

## 1. Introduction

In recent years, the increasing demand for disaster relief humanoids has required more intensive knowledge of human locomotion, as it is the most vital source to inspire the movement and the working posture of humanoids. This demand also benefits the development of rehabilitation orthosis and prosthesis that closely concerns with human-robot interaction.

Over the past decades, humanoid robots have been perceived as one of the most important research fields that receive the interest of scientists and researchers all around the world. So far, humanoid robots have been assisting humans in various activities such as military, medical, education, rescue, exploration, research, etc. (Saputra *et al.*, 2016; Villarreal, Quintero and Gregg, 2016; Panzirsch *et al.*, 2017; Yang *et al.*, 2020). Of all the studies on various aspects of a humanoid, the research on the walking process appears to be the most essential and needs improving considerably. With the goal of perfectly mimicking the human walking gait, the research on efficiently controlling the joints of the humanoid's lower limb must be taken into consideration.

Traditionally, the control strategies for the walking movement of humanoids consist of two main steps. First, the desired walking pattern for humanoids is predefined with the aim of ensuring the stability of the whole body. Then, a controller is designed to control the joint's rotation angle tracking desired trajectories obtained through the inverse kinematic model (Park, Kim and Oh, 2012). The research on establishing a stable walking pattern for humanoids, in general, has observed some breakthrough developments over the years. Vukobratovic (Vukobratović and Stepanenko, 1972) introduced the concept of zero moment point (ZMP) for the first time in 1972, which was a great contribution for the robotics industry and has led to the development of many impressive biped robots (Jun, Ellenberg and Oh, 2010; Aghbali *et al.*, 2013; Alcaraz-Jiménez, Herrero-Pérez and Martínez-Barberá, 2013). In 2013, S. Kajita et al. proposed the preview control strategy that gradually modifies the COM trajectory of humanoids during the walking process (Kajita *et al.*, 2014). Nowadays, several humanoids of big research groups around the world have been developed with their own walking control algorithm such as ASIMO of HONDA, HRP-3 of AIST. However, the process of controlling joint trajectories to ensure the desired walking gait still faces many difficulties. In fact, those traditional control strategies depend on explicit knowledge about kinematics and dynamics of humanoids as well as a well-perceived operating environment. Since the dynamics of humanoids usually introduces nonlinear factors as well as parameter uncertainties, both internal and external, the usages of classical controller like PID or compute torque control (CTC) seem to be inappropriate. To overcome this issue, a robust control algorithm must be adopted. The model reference adaptive controller (MRAC) is well suited for the humanoid-control problem due to its capability to maintain the robustness of the whole system even with the presence of bounded disturbance or variations of some internal parameters.

Besides, the selection of actuators for driving the humanoid's joints is also an important aspect that requires to be considered. The latest designs of humanoid locomotives focus on utilizing compliant mechanisms that can support a wide range of motion while costing low energy. The basic idea to construct a compliant actuator is the addition of spring and damping element to the traditional stiff actuators in order to achieve the desired passive characteristic of a biped. Series Elastic Actuators (SEA) (Pratt and Williamson, 1995) is a notable representative of this doctrine, which was applied in various humanoid platforms competing in the DARPA trial (Knabe, 2015; Knabe *et al.*, 2015; Radford *et al.*, 2015). Overall, the compliant actuators can store and release energy similar to human muscle, enabling humanoids to improve power efficiency and stabilize ground contact (Torricelli *et al.*, 2016).

There were many approaches to the SEA's control strategy. Most of the existing studies used a purely traditional PID controller (or as well P, PI, PD) (Sensinger and Weir, 2006; Ragonesi *et al.*, 2011; Calanca, Muradore and Fiorini, 2017) with relatively low gain to protect the system from instability. Some research deals with the backlash and delay phenomenon in the SEA's design by utilizing an enhanced PID controller in combination with position feedback together with lead-lag compensator and low-pass filter (Wyeth, 2006; Lagoda *et al.*, 2010; Tilburg, 2010). Moreover, in (Kong, Bae and Tomizuka, 2012; C and S, 2019), besides a PID controller, a disturbance observer (DOB) is subjected to nonlinear and undefined disturbances by nominalizing dynamics of target plants into the nominal plant designed in DOB in order to reject the negative effect of these unmeasurable factors. Another solution to deal with disturbance is proposed in (Amaral and Siqueira, 2012) with the usage of $\mathcal{H}_\infty$ criteria to design a robust force controller with consecutively changing gain PID.

In our previous studies (Truong *et al.*, 2020a, 2020b), the lower limb of a commercial humanoid robot called UXA-90, motivated by linear SEAs equipped for each joint and controlled by a traditional PID, can only well perform in a predefined environment, but it suffered some significant vibration for the whole system when impacted by unpredicted external forces. One of our recent research (Lanh *et al.*, 2021) proposed a sliding mode controller to deal with the nonlinearity and predictable disturbance. However, the influence of parameter uncertainties on the stability of the lower limb system is still a challenge and yet has not been solved.

In this paper, research on applying a robust controller for a humanoid's lower limb is implemented by validating the performance of a single leg of a humanoid. First, an overall dynamic model for a single leg is obtained. Secondly, the dynamic model of the SEA is given and analyzed to identify its internal parameters. Then a model reference adaptive controller (MRAC) combined with a back-stepping algorithm is respectively designed based on the dynamic characteristics of the entire system. Finally, some simulations and experiments are carried out to evaluate the appropriation and effectiveness of the proposed controller.

## 2. System Modelling

### *2.1 Lower-limb system modelling*

In Figure 1, an overall model of a single leg of the UXA-90 humanoid is illustrated which can be separated into two main parts: the lower limb and the SEA. The lower limb is demonstrated by an L-shape link rotating about point E. To facilitate further calculation, the total mass distributed on the L-shape link is simplified to one point ($m$) and the SEA's end effector is connected to point C of the L-shape link. The prismatic movement and linear force of the SEAs result in the rotation movement and external torque acting on the L-shape link.

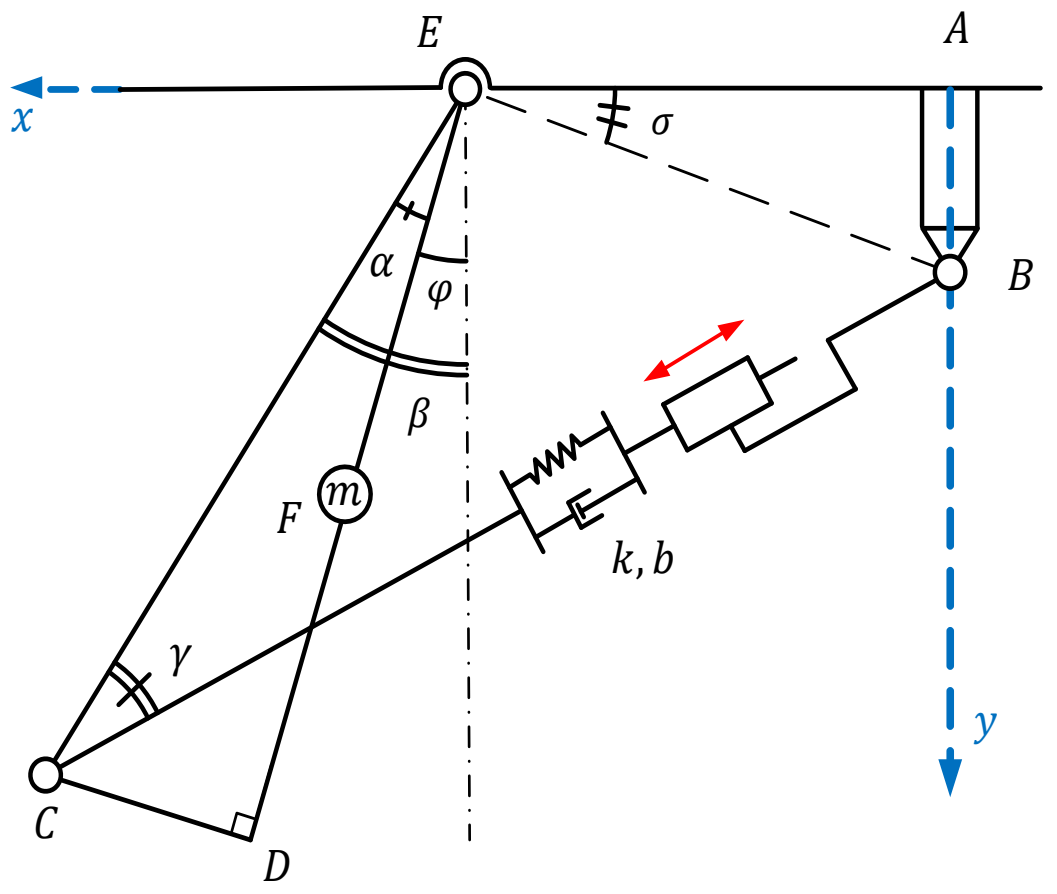

**Figure 1.** The dynamic model of the overall system

The geometric parameters in the proposed dynamic model are denoted as $CD = d_1, DF = d_2, FE = d_3, EA = d_4, AB = d_5, CE = d_6, EB = d_7$.

Define $\tau_{SEA}$ as the torque generated by the SEA's output force $F_{SEA}$ at point C, $\varphi$ is the absolute rotation angle of the L-shape link, and $\beta$ is the angle formed by CE and $y$ axis which related to $\varphi$ by the formula of $\beta = \alpha + \varphi$.

Applying the second Newton's Law for the rotation movement of total mass $m$ about point E, the dynamic of the system described in Figure 1 is given:

$$md_3^2\ddot{\varphi} = -D\dot{\varphi} - mgd_3 \sin\varphi + \tau_{SEA} + \tau_D \tag{1}$$

in which $D$ is the damping coefficient of the joint, $\tau_D$ is the system's disturbance, and $g$ is the gravitational acceleration.

The value of $\tau_{SEA}$ depends on both output force $F_{SEA}$ and overall length of the SEA, which is calculated as:

$$\tau_{SEA} = F_{SEA} d_6 sin\,(\gamma) \tag{2}$$

Let $L_{SEA}(\varphi)$ be the total length of the SEA, the angle of $\gamma$ can be calculated by:

$$sin(\gamma) = \frac{d_7}{L_{SEA}(\varphi)} sin\left(\varphi + \alpha + \frac{\pi}{2} - \sigma\right) \tag{3}$$

By defining $\delta(\varphi) \triangleq \varphi + \alpha + \frac{\pi}{2} - \sigma$ and combining with Eq. (3), Eq. (2) becomes:

$$\tau_{SEA} = F_{SEA} \frac{d_6 d_7\, sin[\delta(\varphi)]}{L_{SEA}(\varphi)} \tag{4}$$

where $\alpha$, $\sigma$ are the fixed angles in the lower limb model that can be determined by utilizing some simple geometric relations.

Hence, by substituting Eq. (4) into Eq. (1), the lower limb dynamics equation can be rewritten as:

$$md_3^2\ddot{\varphi} = -D\dot{\varphi} - mgd_3\, sin\,\varphi + \tau_D + F_{SEA}\frac{d_6 d_7\, sin\,\delta(\varphi)}{L_{SEA}(\varphi)} \tag{5}$$

Since $D$, $\tau_D$, and other geometric parameters are introduced as system's uncertainties, to achieve the control objective, an adaptive back-stepping control law is implemented to guarantee the robustness of the overall system in the presence of uncertain parameters and external disturbances.

*2.2 SEA mechanical modelling*

First, from the mechanical perspective, the dynamic characteristic of the SEA is obtained by applying Newton's second law for the movement of point C along the SEA's direction:

$$m_C \ddot{x}_C = F_{SEA} - F_R \tag{6}$$

where $m_C$ is a pseudo-mass that represents the total mass $m$ at point F but generating the same moment inertia about point E, $x_C$ is the displacements of the SEA's end-effector along the SEA, and $F_R$ is the reaction force generated by the total mass $m$ on point C.

The output force of the SEA includes the deformation of the spring and affects the damping element. However, this damping factor is simplified and included in the lower limb's system damping coefficient $D$. Therefore, the force $F_{SEA}$ can be determined as:

$$F_{SEA} = -k(x_C - x_0) \tag{7}$$

where $k$ is the stiffness coefficient of the spring equipped for the SEA, and $x_0$ is the displacement of the screw's nut.

Let $\tau_R$ be the corresponding reaction torque caused by $F_R$, which can be calculated as:

$$\tau_R = mgd_3 \sin\varphi \tag{8}$$

Then, the reaction force acting along the SEA at point C is determined as:

$$F_R = \frac{\tau_R}{d_6 \sin\gamma} = \frac{mgd_3 \sin(\varphi)}{d_6 d_7 \sin\delta(\varphi)} L_{SEA}(\varphi) \tag{9}$$

The total length of SEA can be obtained by solving the geometric relationship of the SEA's system:

$$L_{SEA}(\varphi) = \sqrt{d_4^2 + d_5^2 + d_6^2 + 2d_6(d_4 \sin\beta - d_5 \cos\beta)} \tag{10}$$

By substituting Eq. (7) and Eq. (9) into Eq. (6), it yields:

$$m_C \ddot{x}_C = -k(x_C - x_0) - \frac{mgd_3 \sin(\varphi)}{d_6 d_7 \sin\delta(\varphi)} L_{SEA}(\varphi) \tag{11}$$

*2.3 SEA electrical modelling*

According to Kirchhoff's circuit law, the dynamic modeling of the armature circuit of an ideal DC motor can be expressed by:

$$V_{IN} = IR + V_{EMF} + L\frac{dI}{dt} = IR + K_{EMF}\omega_M + \frac{L}{K_T}\frac{dT_M}{dt} \tag{12}$$

where $V_{IN}$ is the DC motor's applied voltage, $I$ is the armature current, $R$ is the motor's internal resistance, $L$ is the inductance, $K_{EMF}$ is the back-EMF constant, $K_T$ is the torque constant, and $T_M$ is the motor's electrical torque.

By applying Newton - Euler equation for the motor's rotor, the dynamics of the rigid part can be obtained as:

$$J_M \dot{\omega}_M = T_M - \frac{1}{n\eta_1}\left(J_S \dot{\omega}_S + \frac{l}{2\pi\eta_2}(m_0 \dot{v}_0 + F_L)\right) - B_M \omega_M \tag{13}$$

where $J_M$, $\omega_M$ and $J_s, \omega_s$ are the moment of inertia, the angular velocity of the motor and the ball screw respectively; $m_0$, $v_0$ are respectively the mass and velocity of the ball screw's nut together with the spring base; $B_M$ is the viscous friction coefficient; $n$ is the gearbox transmission ratio; $l$ is the lead of ball screw; $\eta_1$, $\eta_2$ are the efficiency coefficients of the gearbox, the ball screw, respectively; and $F_L$ is the reaction force from the load.

The equivalent moment of inertia of the system $J_{eq}$, which includes the moment of inertia of the motor, ball screw, ball screw's nut, and spring base, is given by:

$$J_{eq} = J_M + \frac{1}{n^2\eta_1} J_s + \frac{l^2}{4\pi^2 n^2 \eta_1 \eta_2} m_0 \tag{14}$$

Substituting Eq. (14) into Eq. (13), and recall that $\omega_M = v_0 \frac{2\pi n}{l}$, the mechanical equation of the rigid part can be represented as:

$$J_{eq}\frac{2\pi n}{l}\dot{v}_0 = T_M - \frac{l}{2\pi n \eta_1 \eta_2} F_L - B_M v_0 \frac{2\pi n}{l} \tag{15}$$

By combining Eq. (15) and Eq. (12), the overall dynamics equation of the SEA is determined as:

$$V_{IN} - \frac{RT_L - L\dot{T}_L}{K_T} = \frac{2\pi n}{lK_T}\left(LJ_{eq}\ddot{v}_0 + \left(RJ_{eq} + LB_M\right)\dot{v}_0 + (B_M R + K_{EMF}K_T)v_0\right) \tag{16}$$

Define $U_v^*$ as a virtual control input, which is expressed as follow:

$$U_v^* \triangleq V_{IN} - \frac{RT_L - L\dot{T}_L}{K_T} \quad (17)$$

Substituting Eq. (17) into Eq. (16), it yields:

$$U_v^* = \frac{2\pi n}{lK_T}\left(LJ_{eq}\ddot{v}_0 + \left(RJ_{eq} + LB_M\right)\dot{v}_0 + (B_M R + K_{EMF}K_T)v_0\right) \quad (18)$$

By using the System Identification toolbox and Simulink toolbox of MATLAB, the internal parameters of the motor and the SEA are identified as $R = 5{,}56\ (\Omega), B = 16{,}5 \times 10^{-5}\ \left(\frac{Nms}{rad}\right), J_M = 1{,}57 \times 10^{-4}\ (Kgm^2)$, $K_T = K_{EMF} = 0{,}202\ \left(\frac{Nm}{A}\right), J_{eq} = 1{,}574 \times 10^{-4} (Kgm^2)$.

Then, substituting the internal parameters into Eq. (18) and neglecting the small value of the coefficient of $\ddot{v}_0$, it yields:

$$U_v^* = 5{,}68\dot{v}_0 + 270v_0 \quad (19)$$

Let assign $U_v = U_v^*/5.68$, and replace $v_0 = \dot{x}_0$, Eq. (19) becomes:

$$U_v = \ddot{x}_0 + 47{,}535\dot{x}_0 \quad (20)$$

Combining Eq. (11) and Eq. (20), the mechanical-electrical dynamic equation of the SEA can be deduced as:

$$(\ddot{x}_C - \ddot{x}_0) + \frac{k}{m_C}(x_C - x_0) + 47{,}535(\dot{x}_c - \dot{x}_0) = -\frac{F_R}{m_C} - U_v + 47{,}535\ \dot{x}_C \quad (21)$$

By defining: $\Delta \triangleq x_C - x_0$, $\omega \triangleq \sqrt{\frac{k}{m}}$, $U_{eq} \triangleq 47{,}535\dot{X}_C - U_v$, and let $\zeta$ be the coefficient of $\dot{x}_C$ ($\zeta = 47{,}535$), Eq. (21) becomes:

$$\ddot{\Delta} + \zeta\dot{\Delta} + \omega^2\Delta = U_{eq} - \frac{F_R}{m_C} \quad (22)$$

*2.4 State-space modeling*

The moment generated by the SEAs is proportional to the deformation of the inner spring that is determined by:

$$\tau_{SEA} = \Delta\frac{-kd_6d_7 \sin\delta(\varphi)}{L_{SEA}(\varphi)} \quad (23)$$

Then it can be obtained:

$$\Delta = \tau_{SEA}\frac{-L_{SEA}(\varphi)}{kd_6d_7 \sin\delta(\varphi)} = \tau_{SEA}G(\varphi) \quad (24)$$

Taking the first and second time-derivative of the Eq. (24), it yields:

$$\dot{\Delta} = \dot{\tau}_{SEA}G(\varphi) + \tau_{SEA}\dot{G}(\varphi) \quad (25)$$

$$\ddot{\Delta} = \ddot{\tau}_{SEA}G(\varphi) + 2\dot{\tau}_{SEA}\dot{G}(\varphi) + \tau_{SEA}\ddot{G}(\varphi) \quad (26)$$

By substituting Eq. (24), Eq. (25), Eq. (26) into Eq. (22), it yields:

$$\ddot{\tau}_{SEA} = \frac{1}{G(\varphi)}[-2\dot{\tau}_{SEA}\dot{G}(\varphi) - \tau_{SEA}\ddot{G}(\varphi) - \omega^2\tau_{SEA}G(\varphi) - \zeta\left(\dot{\tau}_{SEA}G(\varphi) + \tau_{SEA}\dot{G}(\varphi)\right) - \frac{gd_3 \sin(\varphi)}{d_6 \sin(\gamma)}] + \frac{1}{G(\varphi)}U_{eq} \quad (27)$$

To sum it up, the overall lower limb powered by the SEA system is described by the following equations:

$$\ddot{\varphi} = -\frac{D}{md_3^2}\dot{\varphi} - \frac{g}{d_3}\sin\varphi + \frac{\tau_D}{md_3^2} + \frac{\tau_{SEA}}{md_3^2} \quad (28)$$

$$\ddot{\tau}_{SEA} = \frac{1}{G(\varphi)}\{-\dot{\tau}_{SEA}\left(2\dot{G}(\varphi) + \zeta G(\varphi)\right) - \tau_{SEA}\left(\ddot{G}(\varphi) + \omega^2 G(\varphi) + \zeta\dot{G}(\varphi)\right) - \frac{gd_3\, sin(\varphi)}{d_6\, sin(\gamma)}\} + \frac{1}{G(\varphi)}U_{eq}$$

Then, the state variables are defined as follow:

$$\begin{cases} x_1 = \varphi \\ \dot{x}_1 = x_2 \\ \dot{x}_2 = -\dfrac{D}{md_3^2}x_2 - \dfrac{g}{d_3}\sin(x_1) + \dfrac{\tau_D}{md_3^2} + \dfrac{\tau_{SEA}}{md_3^2} \end{cases} \tag{29}$$

$$\begin{cases} z_1 = \tau_{SEA} \\ \dot{z}_1 = z_2 \\ \dot{z}_2 = \dfrac{1}{G(x_1)}\left(-z_2\left(2\dot{G}(x_1) + \zeta G(x_1)\right) - z_1\left(\ddot{G}(x_1) + \omega^2 G(x_1) + \zeta\dot{G}(x_1)\right) - \dfrac{gd_3\, sin(x_1)}{d_6\, sin(\gamma)}\right) + \dfrac{1}{G(x_1)}U_{eq} \end{cases} \tag{30}$$

By utilizing the expression in Eq. (29) and Eq. (30), the overall lower limb powered by SEA dynamic equations become:

$$\begin{bmatrix} \dot{x}_1 \\ \dot{x}_2 \end{bmatrix} = \begin{bmatrix} 0 & 1 \\ 0 & -\dfrac{D}{md_3^2} \end{bmatrix}\begin{bmatrix} x_1 \\ x_2 \end{bmatrix} + \frac{1}{md_3^2}\begin{bmatrix} 0 \\ -mgd_3\sin(x_1) + \tau_D + z_1 \end{bmatrix} \tag{31}$$

$$\begin{bmatrix} \dot{z}_1 \\ \dot{z}_2 \end{bmatrix} = \begin{bmatrix} 0 & 1 \\ -\dfrac{\ddot{G}(x_1) + \omega^2 G(x_1) + \zeta\dot{G}(x_1)}{G(x_1)} & -\dfrac{2\dot{G}(x_1) + \zeta G(x_1)}{G(x_1)} \end{bmatrix}\begin{bmatrix} z_1 \\ z_2 \end{bmatrix} + \frac{1}{G(x_1)}\begin{bmatrix} 0 \\ -\dfrac{gd_3\sin(x_1)}{d_6\sin(\gamma)} + U_{eq} \end{bmatrix} \tag{32}$$

## 3. Controller Design

### *3.1 Model reference adaptive torque control*

By defining the following matrices: $\boldsymbol{X} \triangleq [x_1 \quad x_2]^T, \boldsymbol{A} \triangleq \begin{bmatrix} 0 & 1 \\ 0 & \frac{-D}{md_3^2} \end{bmatrix}, \boldsymbol{B} \triangleq [0 \quad 1]^T, \Lambda \triangleq \frac{1}{md_3^2}$ , $v_x \triangleq \tau_{SEA}$ $f \triangleq [mgd_3 \quad -\tau_D]\begin{bmatrix} sin(x_1) \\ 1 \end{bmatrix} = \boldsymbol{\Theta}^{\boldsymbol{T}}\boldsymbol{\Phi}(\boldsymbol{X})$, Eq. (31) becomes:

$$\dot{\boldsymbol{X}} = \boldsymbol{AX} + \boldsymbol{B}\Lambda(v_x - f) \tag{33}$$

Next, a stable reference model is defined as:

$$\dot{\boldsymbol{X}}_{\boldsymbol{m}} = \boldsymbol{A}_{\boldsymbol{m}}\boldsymbol{X}_{\boldsymbol{m}} + \boldsymbol{B}_{\boldsymbol{m}}r \tag{34}$$

where $\boldsymbol{A}_{\boldsymbol{m}}$ is $2 \times 2$ Hurwitz matrix, $\boldsymbol{B}_{\boldsymbol{m}} \in R^{2\times 1}$, $r$ is the referenced signal known as the desired absolute rotation angle of the L-shape link.

To achieve the goal that $\lim_{t\to\infty}||\boldsymbol{X} - \boldsymbol{X}_{\boldsymbol{m}}|| = 0$, the control signal $u$ is defined as:

$$v_x = \widehat{\boldsymbol{K}}_{\boldsymbol{x}}^{\boldsymbol{T}}\boldsymbol{X} + \widehat{K}_r r + \widehat{\boldsymbol{\Theta}}^{\boldsymbol{T}}\boldsymbol{\Phi}(\boldsymbol{X}) \tag{35}$$

in which, $\widehat{\boldsymbol{K}}_{\boldsymbol{x}} \in R^{2\times 1}$, $\widehat{K}_r \in R$ are the estimated controller gains corresponding to the variation of $\boldsymbol{X}$ and $r$ respectively.

Using Eq. (35), Eq. (33) becomes:

$$\dot{\boldsymbol{X}} = (\boldsymbol{A} + \boldsymbol{B}\Lambda\widehat{\boldsymbol{K}}_{\boldsymbol{x}}^{\boldsymbol{T}})\boldsymbol{X} + \boldsymbol{B}\Lambda\left(\widehat{K}_r r + \left(\widehat{\boldsymbol{\Theta}} - \boldsymbol{\Theta}\right)^{\boldsymbol{T}}\boldsymbol{\Phi}(\boldsymbol{X})\right) \tag{36}$$

Assume the existence of the ideal controller gains, noted as $\boldsymbol{K}_{\boldsymbol{x}}$ and $K_r$ , so that:

$$\begin{cases} \boldsymbol{A} + \boldsymbol{B}\Lambda\boldsymbol{K}_{\boldsymbol{x}}^{\boldsymbol{T}} = \boldsymbol{A}_{\boldsymbol{m}} \\ \boldsymbol{B}\Lambda K_r = \boldsymbol{B}_{\boldsymbol{m}} \end{cases} \tag{37}$$

which leads to the following expression:

$$\begin{cases} \boldsymbol{A} + \boldsymbol{B}\Lambda\widehat{\boldsymbol{K}}_{\boldsymbol{x}}^{\boldsymbol{T}} - \boldsymbol{A}_{\boldsymbol{M}} = \boldsymbol{B}\Lambda\left(\widehat{\boldsymbol{K}}_{\boldsymbol{x}} - \boldsymbol{K}_{\boldsymbol{x}}\right)^{\boldsymbol{T}} = \boldsymbol{B}\Lambda(\boldsymbol{\Delta K}_{\boldsymbol{x}}^{\boldsymbol{T}}) \\ \boldsymbol{B}\Lambda\widehat{K}_r - \boldsymbol{B}_{\boldsymbol{M}} = \boldsymbol{B}\Lambda\left(\widehat{K}_r - K_r\right) = \boldsymbol{B}\Lambda(\Delta K_r) \end{cases} \tag{38}$$

Define the tracking error: $\boldsymbol{e}(t) = \boldsymbol{X} - \boldsymbol{X}_{\boldsymbol{m}}$

Take the first time-derivative of the tracking error, and it can be obtained as:

$$\begin{aligned}\dot{\boldsymbol{e}}(t) &= \dot{\boldsymbol{X}} - \dot{\boldsymbol{X}}_{\boldsymbol{m}} \\ &= \boldsymbol{A_m e} + \boldsymbol{B}\Lambda\left(\boldsymbol{\Delta K_x^T X} + \Delta K_r r + \boldsymbol{\Delta\Theta^T \Phi}(\boldsymbol{X})\right) \\ &= \boldsymbol{A_m e} + \boldsymbol{B}\Lambda\boldsymbol{\delta}(\boldsymbol{X}, r)\end{aligned} \tag{39}$$

Next, let $Q$ be an arbitrary symmetric positive definite matrix. The algebraic Lyapunov equation has a unique solution $P$, which means:

$$\boldsymbol{PA_m} + \boldsymbol{A_m^T P} = -\boldsymbol{Q} \tag{40}$$

Then, a Candidate Lyapunov Function (CLF) of the lower limb system is proposed as:

$$V_x = \boldsymbol{e^T P e} + \Lambda\, tr(\boldsymbol{\Delta K_x^T \gamma_x^{-1} \Delta K_x}) + \Lambda\, tr(\Delta K_r^T \gamma_r^{-1} \Delta K_r) + \Lambda\, tr\left(\boldsymbol{\Delta\Theta^T \gamma_\theta^{-1} \Delta K_\theta}\right) \tag{41}$$

where $\boldsymbol{\gamma_x}, \boldsymbol{\gamma_\theta} \in R^{2\times 2}$ are symmetric positive definite matrices, $\gamma_r > 0 \in R$. The matrices of $\boldsymbol{\gamma_x}, \boldsymbol{\gamma_\theta}$ and positive number of $\gamma_r$ are called adaptation gains.

Take the time-derivative of the CLF:

$$\begin{aligned}\dot{V}_x &= [\boldsymbol{\dot{e}^T P e} + \boldsymbol{e^T P \dot{e}}] + \left[2\Lambda\, tr\left(\boldsymbol{\Delta K_x^T \gamma_x^{-1} \dot{\hat{K}}_x}\right)\right] + \left[2\Lambda\, tr\left(\Delta K_r \gamma_r^{-1} \dot{\hat{K}}_r\right)\right] + \left[2\Lambda\, tr\left(\boldsymbol{\Delta\Theta^T \gamma_\theta^{-1} \dot{\hat{\Theta}}}\right)\right] \\ &= \left[\left(\boldsymbol{A_m e} + \boldsymbol{B}\Lambda\boldsymbol{\delta}(\boldsymbol{X}, r)\right)^{\boldsymbol{T}} \boldsymbol{Pe} + \boldsymbol{e^T P}\left(\boldsymbol{A_m e} + \boldsymbol{B}\Lambda\boldsymbol{\delta}(\boldsymbol{X}, r)\right)\right] \\ &\quad +2\Lambda\left[tr\left(\boldsymbol{\Delta K_x^T \gamma_x^{-1} \dot{\hat{K}}_x}\right) + tr\left(\Delta K_r \gamma_r^{-1} \dot{\hat{K}}_r\right) + tr\left(\boldsymbol{\Delta\Theta^T \gamma_\theta^{-1} \dot{\hat{\Theta}}}\right)\right]\end{aligned} \tag{42}$$

$$\begin{aligned} &= [\boldsymbol{e^T}(\boldsymbol{A_m^T P} + \boldsymbol{PA_m})\boldsymbol{e} + 2\boldsymbol{e^T PB}\Lambda\boldsymbol{\delta}(\boldsymbol{X}, r)] \\ &\quad +2\Lambda\left(tr\left(\boldsymbol{\Delta K_x^T \gamma_x^{-1} \dot{\hat{K}}_x}\right) + tr\left(\Delta K_r \gamma_r^{-1} \dot{\hat{K}}_r\right) + tr\left(\boldsymbol{\Delta\Theta^T \gamma_\theta^{-1} \dot{\hat{\Theta}}}\right)\right) \\ &= -\boldsymbol{e^T Q e} \\ &\quad +2\Lambda\left(\boldsymbol{e^T PB\Delta K_x^T X} + tr\left(\boldsymbol{\Delta K_x^T \gamma_x^{-1} \dot{\hat{K}}_x}\right)\right) \\ &\quad +2\Lambda\left(\boldsymbol{e^T PB\Delta K_r r} + tr\left(\Delta K_r \gamma_r^{-1} \dot{\hat{K}}_r\right)\right) \\ &\quad +2\Lambda\left(\boldsymbol{e^T PB\Delta\Theta^T \Phi}(\boldsymbol{X}) + tr\left(\boldsymbol{\Delta\Theta^T \gamma_\theta^{-1} \dot{\hat{\Theta}}}\right)\right)\end{aligned} \tag{43}$$

By applying the $trace$ function's property for vector: $\forall \boldsymbol{a}, \boldsymbol{b} \in R^{n\times 1}\; tr(\boldsymbol{ba^T}) = \boldsymbol{a^T b}$, Eq. (43) becomes:

$$\begin{aligned}\dot{V}_x = -\boldsymbol{e^T Q e} + 2\Lambda\Big(&tr\left(\boldsymbol{\Delta K_x^T}\left\{\boldsymbol{\gamma_x^{-1} \dot{\hat{K}}_x} + \boldsymbol{Xe^T PB}\right\}\right) + tr\left(\Delta K_r\left\{\gamma_r^{-1}\dot{\hat{K}}_r + r\boldsymbol{e^T PB}\right\}\right) \\ &+ tr\left(\boldsymbol{\Delta\Theta^T}\left\{\boldsymbol{\gamma_\theta^{-1}\dot{\hat{\Theta}}} + \boldsymbol{\Phi}(\boldsymbol{X})\boldsymbol{e^T PB})\right\}\right)\Big)\end{aligned} \tag{44}$$

The adaptation laws are chosen as:

$$\begin{aligned}\dot{\hat{\boldsymbol{K}}}_x &= -\boldsymbol{\gamma_x X e^T PB} \\ \dot{\hat{K}}_r &= -\gamma_r r\boldsymbol{e^T PB} \\ \dot{\hat{\boldsymbol{\Theta}}} &= -\boldsymbol{\gamma_\theta \Phi}(\boldsymbol{X})\boldsymbol{e^T PB}\end{aligned} \tag{45}$$

By using the adaptation laws of Eq. (45), the time-derivative of the CLF is:

$$\dot{V}_x = -\boldsymbol{e^T Q e} \le \boldsymbol{0} \tag{46}$$

This implies that the tracking error $\boldsymbol{e}$ and $\boldsymbol{\Delta K_x^T}, \Delta K_r, \boldsymbol{\Delta\Theta^T}$ are bounded. By taking the second time-derivative of the CLF, it yields:

$$\ddot{V}_x = -2(\boldsymbol{e}^T \boldsymbol{Q}\, \dot{\boldsymbol{e}}) \tag{47}$$

Recall from Eq. (39), the first time-derivative of tracking error is $\dot{\boldsymbol{e}} = \boldsymbol{A_m e} + \boldsymbol{B}\Lambda\big(\boldsymbol{\Delta K_x^T X} + \Delta K_r r + \boldsymbol{\Delta\Theta^T \Phi(X)}\big)$. The term of $\boldsymbol{e}, \boldsymbol{\Delta K_x^T}, \Delta K_r$ and $\boldsymbol{\Delta\Theta^T}$ are proven to be bounded. The referenced signal $r$ includes predefined values of the desired rotation angle, so that it is undoubtedly bounded. Meanwhile, since $\boldsymbol{A_m}$ is a Hurwitz matrix, the reference model is ensured to be stable; and with a bounded input signal of $r$, the output response $\boldsymbol{X_m}$ of the reference model is also bounded. Because $\boldsymbol{e}$ and $\boldsymbol{X_m}$ are already bounded, the lower limb system's response $\boldsymbol{X}$ and $\boldsymbol{\Phi}(\boldsymbol{X})$, a harmonic function of $\boldsymbol{X}$, are consequently bounded. For the aforementioned reasons, $\dot{\boldsymbol{e}}$ is showed to be bounded, and so that $\ddot{V}_x$ is bounded. The fact that $\ddot{V}_x$ is bounded means $\dot{V}_x$ is a uniform continuity function. According to Barbalat's lemma, this leads to the conclusion that $\lim_{t\to\infty}(\dot{V}_x) = 0$ and the lower limb system is stable.

*3.2 Back-stepping algorithm*

To apply the back-stepping approach control, it would be better that the Eq. (33) be expressed in the following form:

$$\dot{\boldsymbol{X}} = \boldsymbol{f_x}(\boldsymbol{X}) + \boldsymbol{g_x}(\boldsymbol{X}) v_x \tag{48}$$

where:

$$\boldsymbol{f_x}(\boldsymbol{X}) \triangleq \begin{bmatrix} x_2 \\ -\dfrac{g}{d_3}\sin(x_1) + \dfrac{\tau_D}{md_3^2} - \dfrac{Bx_2}{md_3^2} \end{bmatrix}$$

$$\boldsymbol{g_x}(\boldsymbol{X}) \triangleq \begin{bmatrix} 0 \\ \dfrac{1}{md_3^2} \end{bmatrix}$$

In the previous section, this system is ensured to be stable with the Lyapunov function $V_x$ in Eq. (41) and adaptation laws in Eq. (45). Consider the first augmented system:

$$\begin{cases} \dot{\boldsymbol{X}} = \boldsymbol{f_x}(\boldsymbol{X}) + \boldsymbol{g_x}(\boldsymbol{X}) z_1 \\ \dot{z}_1 = f_1(\boldsymbol{X_1}) + g_1(\boldsymbol{X_1}) v_1 \end{cases} \tag{49}$$

in which, $\boldsymbol{X_1} \triangleq [x_1 \quad x_2 \quad z_1]^T$, $f_1(\boldsymbol{X_1}) \triangleq 0$, $g_1(\boldsymbol{X_1}) \triangleq 1$

In the previous section, an adaptive control law $v_x$ is chosen to ensure the stability of the lower limb system. Define a pseudo-control variable $v_1$. The control goal is to make $z_1$ follow the desired value of $v_x$. The augmented Lyapunov function is chosen as:

$$V_1 = V_x + \frac{1}{2}(z_1 - v_x)^2 \tag{50}$$

Using the expression of $\dot{V}_x$ in Eq. (42), the first time-derivative of Eq. (50) is given by:

$$\begin{aligned} \dot{V}_1 = {} & [\boldsymbol{e^T}(\boldsymbol{A_m^T P} + \boldsymbol{PA_m})\boldsymbol{e} + 2\boldsymbol{e^T PB}\Lambda\boldsymbol{\delta}(\boldsymbol{X}, r)] \\ & + 2\Lambda\left(tr\left(\boldsymbol{\Delta K_x^T \gamma_x^{-1} \dot{\hat{K}}_x}\right) + tr\left(\Delta K_r \gamma_r^{-1} \dot{\hat{K}}_r\right) + tr\left(\boldsymbol{\Delta\Theta^T \gamma_\theta^{-1} \dot{\hat{\Theta}}}\right)\right) + (z_1 - v_x)(\dot{z}_1 - \dot{v}_x) \end{aligned} \tag{51}$$

In order to simplify Eq. (51), the following terms are defined as:

$$\begin{aligned} \psi &\triangleq \boldsymbol{K_x^T X} + K_r r + \boldsymbol{\Theta^T \Phi}(\boldsymbol{X}) \\ \varepsilon &\triangleq 2\Lambda\left(tr\left(\boldsymbol{\Delta K_x^T \gamma_x^{-1} \dot{\hat{K}}_x}\right) + tr\left(\Delta K_r \gamma_r^{-1} \dot{\hat{K}}_r\right) + tr\left(\boldsymbol{\Delta\Theta^T \gamma_\theta^{-1} \dot{\hat{\Theta}}}\right)\right) \end{aligned} \tag{52}$$

Then, by using expressions in Eq. (52), Eq. (51) becomes:

$$\begin{aligned} \dot{V}_1 &= [\boldsymbol{e^T}(\boldsymbol{A_m^T P} + \boldsymbol{PA_m})\boldsymbol{e} + 2\boldsymbol{e^T PB}\Lambda(z_1 - \psi)] + \varepsilon + (z_1 - v_x)(z_2 - \dot{v}_x) \\ &= [\boldsymbol{e^T}(\boldsymbol{A_m^T P} + \boldsymbol{PA_m})\boldsymbol{e} + 2\boldsymbol{e^T PB}\Lambda(v_x - \psi)] + \varepsilon + 2\boldsymbol{e^T PB}\Lambda(z_1 - v_x) + (z_1 - v_x)(v_1 - \dot{v}_x) \end{aligned} \tag{53}$$

To stabilize the proposed augmented system, the pseudo-control law $v_1$ is chosen as:

$$v_1 = \dot{v}_x - 2\boldsymbol{e^T PB}\Lambda - k_1(z_1 - v_x) \tag{54}$$

where $v_x = \boldsymbol{\hat{K}_x}^{\boldsymbol{T}}\boldsymbol{X} + \hat{K}_r r + \boldsymbol{\hat{\Theta}^T \Phi}(\boldsymbol{X})$

The second augmented system is chosen as:

$$\begin{cases} \dot{\boldsymbol{X}}_{\boldsymbol{1}} = \boldsymbol{f}_{\boldsymbol{1}}(\boldsymbol{X}_{\boldsymbol{1}}) + \boldsymbol{g}_{\boldsymbol{1}}(\boldsymbol{X}_{\boldsymbol{1}}) z_2 \\ \dot{z}_2 = f_2(\boldsymbol{X}_{\boldsymbol{2}}) + g_2(\boldsymbol{X}_{\boldsymbol{2}}) U_{eq} \end{cases} \tag{55}$$

in which:

$$\boldsymbol{X}_{\boldsymbol{2}} \triangleq [\boldsymbol{X}_{\boldsymbol{1}} \quad z_2]^T$$

$$f_2(\boldsymbol{X}_{\boldsymbol{2}}) \triangleq \frac{1}{G(x_1)}\left(-2z_2\dot{G}(x_1) - z_1\left(\ddot{G}(x_1) + \omega^2 G(x_1)\right) - \frac{gd_3\sin(x_1)}{d_6\sin(\gamma)}\right)$$

$$g_2(\boldsymbol{X}_{\boldsymbol{2}}) \triangleq \frac{1}{G(x_1)}$$

To achieve the goal of making $z_2$ follow the desired value of $v_1$, the ultimate control law $U_{eq}$ is determined by considering the $2^{nd}$ augmented Lyapunov function:

$$V_2 = V_1 + \frac{1}{2}(z_2 - v_1)^2 \tag{56}$$

Taking time derivative of Eq. (56), it yields:

$$\begin{aligned} \dot{V}_2 &= \dot{V}_1 + (z_2 - v_1)(\dot{z}_2 - \dot{v}_1) \\ &= [\boldsymbol{e}^{\boldsymbol{T}}(\boldsymbol{A}_{\boldsymbol{m}}^{\boldsymbol{T}}\boldsymbol{P} + \boldsymbol{P}\boldsymbol{A}_{\boldsymbol{m}})\boldsymbol{e} + 2\boldsymbol{e}^{\boldsymbol{T}}\boldsymbol{P}\boldsymbol{B}\Lambda(v_x - \psi)] + \varepsilon + 2\boldsymbol{e}^{\boldsymbol{T}}\boldsymbol{P}\boldsymbol{B}\Lambda(z_1 - v_x) + (z_1 - v_x)(z_2 - \dot{v}_x) \\ &\quad + (z_2 - v_1)(\dot{z}_2 - \dot{v}_1) \\ &= [\boldsymbol{e}^{\boldsymbol{T}}(\boldsymbol{A}_{\boldsymbol{m}}^{\boldsymbol{T}}\boldsymbol{P} + \boldsymbol{P}\boldsymbol{A}_{\boldsymbol{m}})\boldsymbol{e} + 2\boldsymbol{e}^{\boldsymbol{T}}\boldsymbol{P}\boldsymbol{B}\Lambda(v_x - \psi)] + \varepsilon + 2\boldsymbol{e}^{\boldsymbol{T}}\boldsymbol{P}\boldsymbol{B}\Lambda(z_1 - v_x) + (z_1 - v_x)(v_1 - \dot{v}_x) \\ &\quad + (z_1 - v_x)(z_2 - v_1) + (z_2 - v_1)(f_2(\boldsymbol{X}_{\boldsymbol{2}}) + g_2(\boldsymbol{X}_{\boldsymbol{2}})U_{eq} - \dot{v}_1) \end{aligned} \tag{57}$$

The final control law is the voltage needed to supply for the DC motor, which is:

$$U_{eq} = \frac{1}{g_2(\boldsymbol{X}_{\boldsymbol{2}})}[-f_2(\boldsymbol{X}_{\boldsymbol{2}}) + \dot{v}_1 - k_2(z_2 - v_1) - (z_1 - v_x)] \tag{58}$$

## 4. Simulation and Results

The simulation is carried out to evaluate the capability of the control algorithm. The sampling time for this simulation is 10 $ms$. The initial value of the state variables is: $x_1(0) = 0.2,\ x_2(0) = 0, z_1(0) = 0, z_2(0) = 0$.

The physical and geometric parameters of the system are given in Table I.

**Table I.** Physical and Geometric parameters

| Physical parameters | Value |
|---|---|
| Mass, $m$ (kg) | 2 |
| Viscous friction coefficient, $D$ ($Nms$) | 0.5 |
| Stiffness coefficient, $k$ ($N/m$) | 20000 |
| **Geometric parameters** | **Value** |
| $d_1$ ($m$) | 0.0280 |
| $d_2$ ($m$) | 0.0525 |
| $d_3$ ($m$) | 0.0525 |
| $d_4$ ($m$) | 0.0350 |
| $d_5$ ($m$) | 0.1180 |

The referenced model is chosen as: $\boldsymbol{A_m} = \begin{bmatrix} 0 & 1 \\ -6 & -4 \end{bmatrix}$, $\boldsymbol{B_m} = \begin{bmatrix} 0 \\ 6 \end{bmatrix}$. The symmetric positive-define matrix, which is the unique solution of Lyapunov algebraic function, is $\boldsymbol{P} = \begin{bmatrix} 3/8 & 1/4 \\ 1/4 & 3/16 \end{bmatrix}$. Through the trials, the best adaptation gain matrices and numbers are determined as: $\boldsymbol{\gamma_x} = \begin{bmatrix} 4000 & 0 \\ 0 & 50 \end{bmatrix}$, $\gamma_r = 2000$, $\boldsymbol{\gamma_\theta} = \begin{bmatrix} 50 & 0 \\ 0 & 50 \end{bmatrix}$. The controller gains in the back-stepping algorithm are chosen as: $k_1 = 30$, $k_2 = 10$. The referenced signal used is the rotation angle at the hip joint of humanoid UXA-90 in a walking cycle, which is inferred from (Nguyen, X. T. *et al.*, 2020).

The simulations of the system's response to the proposed model and adaptation gains are carried out to test the stabilization of the system. The results are shown in Figure 2 and Figure 3.

In Figure 2, the desired absolute rotation angle trajectory is expressed by a dashed line, the response of the proposed reference model is represented by a solid line, and a dotted line describes the simulated response of the lower limb system. Meanwhile, Figure 3 shows the tracking error in absolute rotation angle $e_1$ and in angular velocity $e_2$ during a walking cycle of humanoid.

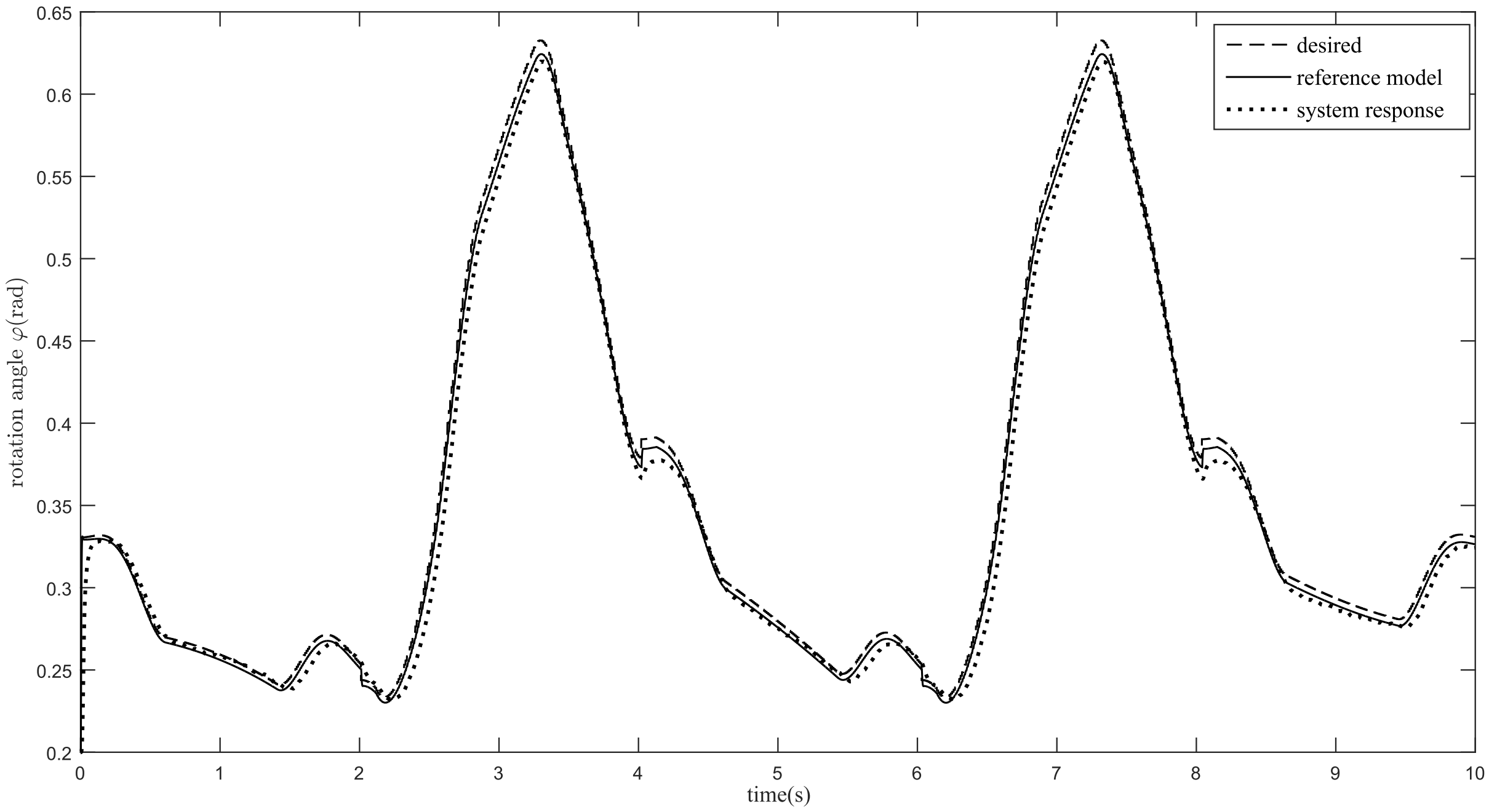

**Figure 2.** Comparison between the desired rotation angle, reference model response and system response simulation

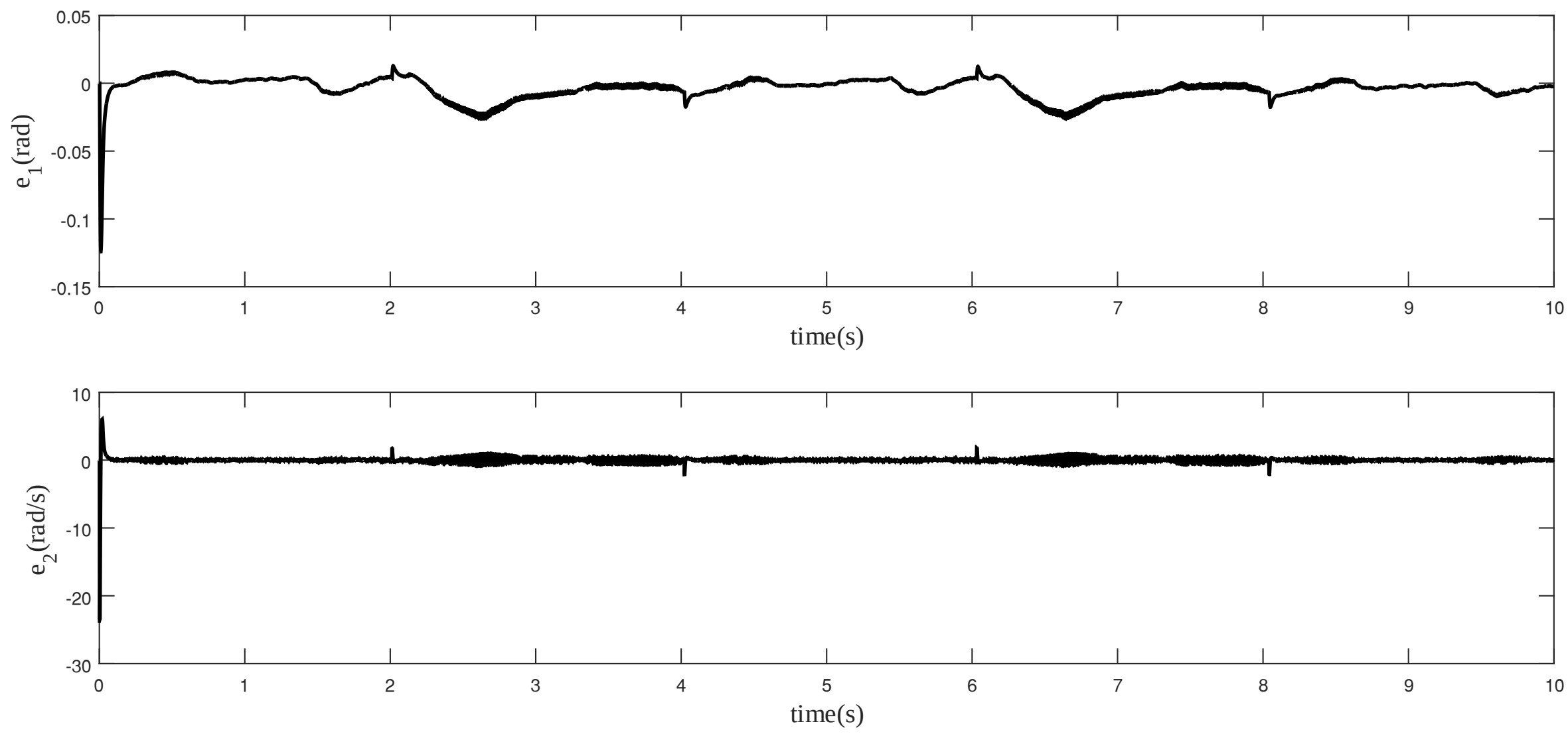

**Figure 3**. The angle tracking error and angular velocity error

It can be seen from Figure 2, the absolute rotation angle of the L-shape link can track the desired angle trajectory with relatively small error. In specific, except for the initial error during the transient phase, the highest absolute angle error $e_1$ is approximate $0.03\ rad$, which is less than $10\%$ compared to the absolute rotation angle at every instant during the whole cycle. The angle tracking error mainly appears at the peak of the graph since it is the moment when the lower limb reverses its rotating direction rapidly. Meanwhile, the angular velocity error $e_2$ rapidly converges to zero but still maintains some oscillations around that steady-state.

Besides, the impact of adaptation gains $\boldsymbol{\gamma_x}$, $\gamma_r$, $\boldsymbol{\gamma_\theta}$ is determined through a trial-and-error approach. Firstly, $\boldsymbol{\gamma_x}$ is responsible for the performance in the transient phase of response, especially the elements of $\gamma_{x11}$ and $\gamma_{x22}$ lying on the main diagonal line of the matrix $\boldsymbol{\gamma_x}$. The increase of $\gamma_{x11}$ would generally reduce the angle tracking error and reduce fluctuations in this phase while the decrease of $\gamma_{x22}$ leads to the shorter transient time. However, a large adaptive gain of $\gamma_{x11}$ may cause overshooting phenomenon while a low adaptive gain of $\gamma_{x22}$ would require much higher torque from the SEA to track the variation of desired angle trajectory. Next, the coefficient of $\gamma_r$ shows that it is significant merits in the steady-state. In specific, a large adaptation gain of $\gamma_r$ strongly reduces the steady-state error, but in return, it also increases the oscillation in angular velocity error due to the consecutive change of $\dot{\varphi}$. Finally, $\boldsymbol{\gamma_\theta}$ plays a key role to retain the robust stability of the lower limb system against external disturbance. A relatively large adaptation gain of $\boldsymbol{\gamma_\theta}$ would effectively eliminate the negative impact of disturbance; however, too much rise of $\boldsymbol{\gamma_\theta}$ may lead to an over sensitive system, which can quickly become unstable.

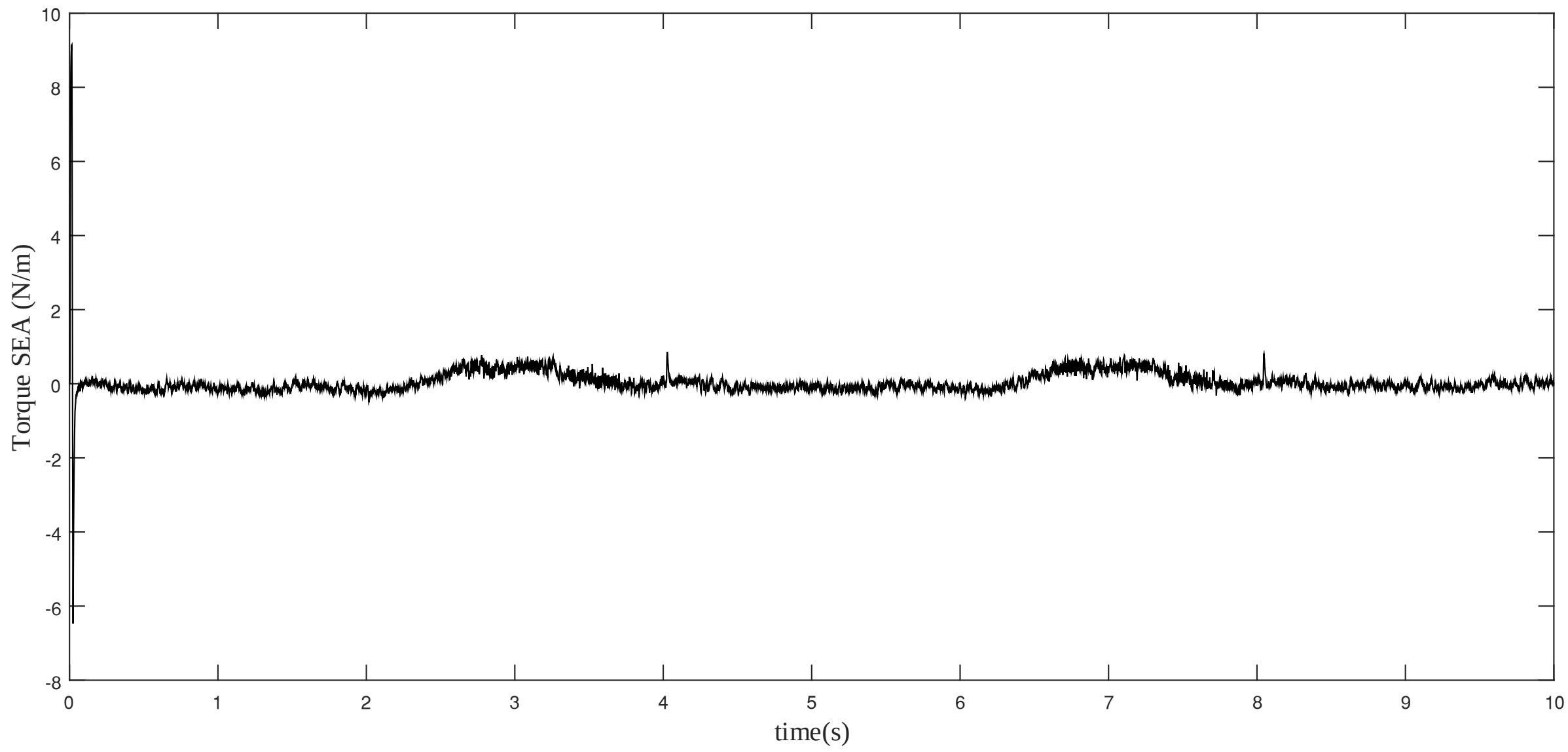

**Figure 4**. Torque generated by the SEA during one walking cycle

Figure 4 illustrates the required torque from the SEA to ensure the well-tracking response for the lower limb system. The highest torque required is approximately 9 $Nm$ in the initial time when the lower limb requires much more power to start moving. However, when the system reaches the steady-state with relatively low angle error, the demand torque remains within the range of $\pm 1.5\ Nm$.

The variation of controller gains is also validated and shown in Figure 5. The controller gains of $K_r$, $K_{x2}$, and $\Theta_2$ rapidly converge to a specific steady-state value. Meanwhile, $K_{x1}$ keeps varying in the same shape with the angle error $e_1$. The reason is that the rotation angle cannot reach the desired trajectory at several instants, which leads to the gradual change of $K_{x1}$. On the other hand, the variation of $\Theta_1$ relies on the external disturbance. Since the disturbance is bounded, the controller gain of $\Theta_1$ is also bounded and slightly oscillates in a certain domain.

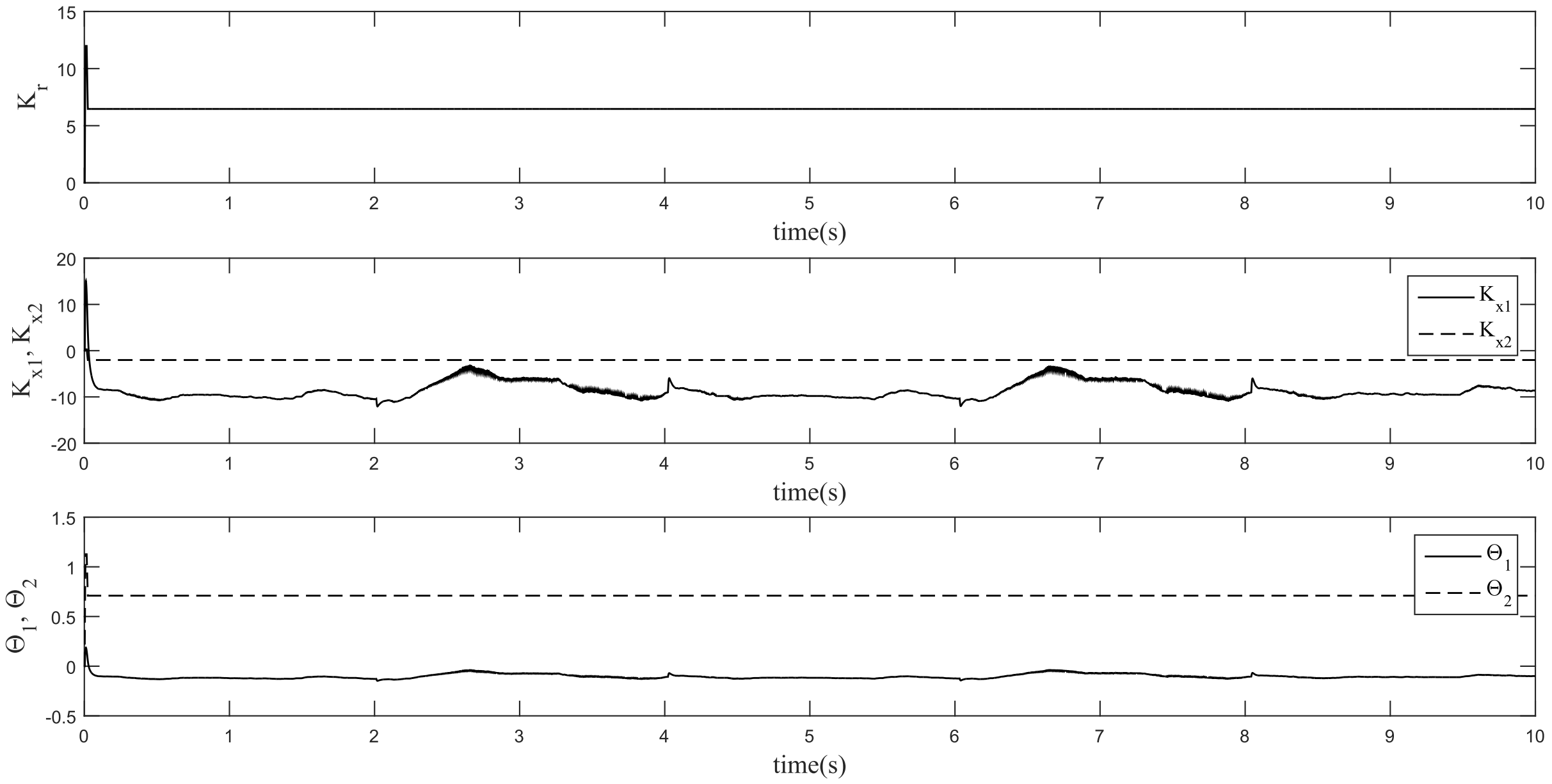

**Figure 5.** The variation of the controller gains $\boldsymbol{K_{x1}}, \boldsymbol{K_{x2}}, \boldsymbol{K_r}, \boldsymbol{\Theta_1}, \boldsymbol{\Theta_2}$

## 5. Conclusion

In conclusion, the model reference adaptive control-based back-stepping control algorithm is proven to be appropriate for this application. The effectiveness and quality of the proposed controller rely on the selections of adaptation gains, which can make both positive and negative impacts on the stability of the system. Moreover, although the simulation showed promising results in this process, the actual response of the system in the practical experiment might be different due to the incorrect estimations of the parameter uncertainties. In the future, several experiment scenarios will be carried out to further evaluate the capability of the proposed controller for controlling the lower limb system.

## Acknowledgements

This research is supported by DCSELab and funded by Vietnam National University HoChiMinh City (VNU-HCM) under grant numbers TX2022-20b-01.

## References

Aghbali, B. *et al.* (2013) 'ZMP trajectory control of a humanoid robot using different controllers based on an offline trajectory generation', in *International Conference on Robotics and Mechatronics, ICRoM 2013*, pp. 530–534. doi: 10.1109/ICROM.2013.6510161.

Alcaraz-Jiménez, J. J., Herrero-Pérez, D. and Martínez-Barberá, H. (2013) 'Robust feedback control of ZMP-based gait for the humanoid robot Nao':, *http://dx.doi.org/10.1177/0278364913487566*. SAGE PublicationsSage UK: London, England, 32(9–10), pp. 1074–1088. doi: 10.1177/0278364913487566.

Amaral, L. M. S. do and Siqueira, A. (2012) 'ROBUST FORCE AND IMPEDANCE CONTROL OF SERIES ELASTIC ACTUATORS'.

C, L. and S, O. (2019) 'Development, Analysis, and Control of Series Elastic Actuator-Driven Robot Leg', *Frontiers in neurorobotics*. Front Neurorobot, 13. doi: 10.3389/FNBOT.2019.00017.

Calanca, A., Muradore, R. and Fiorini, P. (2017) 'Impedance control of series elastic actuators: Passivity and acceleration-based control', *Mechatronics*. Pergamon, 47, pp. 37–48. doi: 10.1016/J.MECHATRONICS.2017.08.010.

Jun, Y., Ellenberg, R. and Oh, P. (2010) 'Realization of miniature humanoid for obstacle avoidance with real-time ZMP preview control used for full-sized humanoid', in *2010 10th IEEE-RAS International Conference on Humanoid Robots, Humanoids 2010*, pp. 46–51. doi: 10.1109/ICHR.2010.5686276.

Kajita, S. *et al.* (2014) *Introduction to Humanoid Robotics, Springer Tracts in Advanced Robotics*. Springer Verlag. doi: 10.1007/978-3-642-54536-8.

Knabe, C. *et al.* (2015) 'Design of a series elastic humanoid for the DARPA Robotics Challenge', in *IEEE-RAS International Conference on Humanoid Robots*. IEEE Computer Society, pp. 738–743. doi:

10.1109/HUMANOIDS.2015.7363452.

Knabe, C. (2015) *Design of Linear Series Elastic Actuators for a Humanoid Robot, undefined*. Virginia Polytechnic Institute and State University.

Kong, K., Bae, J. and Tomizuka, M. (2012) 'A compact rotary series elastic actuator for human assistive systems', *IEEE/ASME Transactions on Mechatronics*, 17(2), pp. 288–297. doi: 10.1109/TMECH.2010.2100046.

Lagoda, C. *et al.* (2010) 'Design of an electric series elastic actuated joint for robotic gait rehabilitation training', in *2010 3rd IEEE RAS and EMBS International Conference on Biomedical Robotics and Biomechatronics, BioRob 2010*, pp. 21–26. doi: 10.1109/BIOROB.2010.5626010.

Lanh, L. A. K. *et al.* (2021) 'Design of robust controller applied for series elastic actuators in controlling humanoid's joint', in *17th International Conference on Intelligent Unmanned Systems (ICIUS 2021)*. The 17th International Conference on Intelligent Unmanned Systems (ICIUS 2021). Available at: https://arxiv.org/abs/2107.08610v1 (Accessed: 21 July 2021).

Panzirsch, M. *et al.* (2017) 'Tele-healthcare with humanoid robots: A user study on the evaluation of force feedback effects', in *2017 IEEE World Haptics Conference, WHC 2017*. Institute of Electrical and Electronics Engineers Inc., pp. 245–250. doi: 10.1109/WHC.2017.7989909.

Park, I.-W., Kim, J.-Y. and Oh, J.-H. (2012) 'Online Walking Pattern Generation and Its Application to a Biped Humanoid Robot — KHR-3 (HUBO)', *http://dx.doi.org/10.1163/156855308X292538*. Taylor & Francis Group , 22(2–3), pp. 159–190. doi: 10.1163/156855308X292538.

Pratt, G. A. and Williamson, M. M. (1995) 'Series elastic actuators', in *IEEE International Conference on Intelligent Robots and Systems*. IEEE, pp. 399–406. doi: 10.1109/IROS.1995.525827.

Radford, N. A. *et al.* (2015) 'Valkyrie: NASA's First Bipedal Humanoid Robot', *Journal of Field Robotics*. John Wiley and Sons Inc., 32(3), pp. 397–419. doi: 10.1002/ROB.21560.

Ragonesi, D. *et al.* (2011) 'Series elastic actuator control of a powered exoskeleton', in *Proceedings of the Annual International Conference of the IEEE Engineering in Medicine and Biology Society, EMBS*, pp. 3515–3518. doi: 10.1109/IEMBS.2011.6090583.

Saputra, A. A. *et al.* (2016) 'Biologically Inspired Control System for 3-D Locomotion of a Humanoid Biped Robot', *IEEE Transactions on Systems, Man, and Cybernetics: Systems*. Institute of Electrical and Electronics Engineers Inc., 46(7), pp. 898–911. doi: 10.1109/TSMC.2015.2497250.

Sensinger, J. W. and Weir, R. F. (2006) 'Unconstrained impedance control using a compact series elastic actuator', in *Proceedings of the 2nd IEEE/ASME International Conference on Mechatronic and Embedded Systems and Applications, MESA 2006*. Institute of Electrical and Electronics Engineers Inc. doi: 10.1109/MESA.2006.296972.

Tilburg, T. M. W. V. (2010) *Control of a 1 degree-of-freedom Series Elastic Actuator on a Humanoid Robot, undefined*. Technische Universiteit Eindhoven.

Torricelli, D. *et al.* (2016) 'Human-like compliant locomotion: state of the art of robotic implementations', *Bioinspiration & biomimetics*. Bioinspir Biomim, 11(5). doi: 10.1088/1748-3190/11/5/051002.

Truong, K. D. *et al.* (2020a) 'A Study on Series Elastic Actuator Applied for Human-Interactive Robot', in *International Conference on Advanced Mechatronic Systems, ICAMechS*. IEEE Computer Society, pp. 7–12. doi: 10.1109/ICAMECHS49982.2020.9310091.

Truong, K. D. *et al.* (2020b) 'Design of Series Elastic Actuator Applied for Humanoid', in *International Conference on Advanced Mechatronic Systems, ICAMechS*. IEEE Computer Society, pp. 23–28. doi: 10.1109/ICAMECHS49982.2020.9310118.

Villarreal, D. J., Quintero, D. and Gregg, R. D. (2016) 'A Perturbation Mechanism for Investigations of Phase-Dependent Behavior in Human Locomotion', *IEEE Access*. Institute of Electrical and Electronics Engineers Inc., 4, pp. 893–904. doi: 10.1109/ACCESS.2016.2535661.

Vukobratović, M. and Stepanenko, J. (1972) 'On the stability of anthropomorphic systems', *Mathematical Biosciences*. Elsevier, 15(1–2), pp. 1–37. doi: 10.1016/0025-5564(72)90061-2.

Wyeth, G. (2006) *Control issues for velocity sourced series elastic actuators, undefined*.

Yang, C. *et al.* (2020) 'Learning Natural Locomotion Behaviors for Humanoid Robots Using Human Bias', *IEEE Robotics and Automation Letters*. Institute of Electrical and Electronics Engineers Inc., 5(2), pp. 2610–2617. doi: 10.1109/LRA.2020.2972879.